\documentclass{article}

\usepackage{arxiv}

\usepackage[utf8]{inputenc} 
\usepackage{float}
\usepackage[T1]{fontenc}    
\usepackage{hyperref}       
\usepackage{url}            
\usepackage{booktabs}       
\usepackage{amsfonts}       
\usepackage{nicefrac}       
\usepackage{microtype}      
\usepackage{lipsum}
\usepackage{fancyhdr}       
\usepackage{graphicx}       
\usepackage{parskip}
\graphicspath{{media/}}     

\title{Control and Coordination of a SWARM of Unmanned Surface Vehicles using Deep Reinforcement Learning in ROS
}

\author{
  Shrudhi R S \\
  Department of Electronics Engineering \\
  Vellore Institute of Technology  \\
  Chennai\\
  \texttt{shrudhi.rs2019@vitstudent.ac.in} \\
  \And
  Sreyash Mohanty \\
  Department of Electronics Engineering \\
  Vellore Institute of Technology \\
  Chennai \\
  \texttt{sreyash.mohanty2019@vitstudent.ac.in} \\
   \And
  Dr. Susan Elias \\
  Dean, Department of Electronics Engineering \\
  Vellore Institute of Technology \\
  Chennai \\
  \texttt{susan.elias@vit.ac.in} \\
}

\begin{document}
\maketitle

\begin{abstract}
An unmanned surface vehicle (USV) can perform complex missions by continuously observing the state of its surroundings and taking action toward a goal. A SWARM of USVs working together can complete missions faster, and more effectively than a single USV alone. In this paper, we propose an autonomous communication model for a swarm of USVs. The goal of this system is to implement a software system using Robot Operating System (ROS) and Gazebo. With the main objective of coordinated task completion, the Markov decision process (MDP) provides a base to formulate a task decision problem to achieve efficient localization and tracking in a highly dynamic water environment. To coordinate multiple USVs performing real-time target tracking, we propose an enhanced multi-agent reinforcement learning approach. Our proposed scheme uses MA-DDPG, or Multi-Agent Deep Deterministic Policy Gradient, an extension of the Deep Deterministic Policy Gradients (DDPG) algorithm that allows for decentralized control of multiple agents in a cooperative environment. MA-DDPG's decentralised control allows each and every agent to make decisions based on its own observations and objectives, which can lead to superior gross performance and improved stability. Additionally, it provides communication and coordination among agents through the use of collective readings and rewards.


\keywords{Reinforcement Learning \and Robot Operating System (ROS) \and Localization and Tracking \and Unmanned Surface Vehicles (USVs) \and Deep Deterministic Policy Gradient (DDPG) \and Multi-Agent DDPG (MA-DDPG)}
\end{abstract}

\section{Introduction}
Unmanned surface vehicles (USVs) have gained increasing attention in recent years, due to their potential impact on various applications, such as oceanographic research, environmental monitoring, and maritime security. The effectiveness of individual USVs in accomplishing complex tasks is limited. To overcome this challenge, swarm-based approaches have emerged as promising solutions. A powerful approach to enable a swarm of USVs is reinforcement learning, a sub-field of machine learning. Using this, the USVs will be capable of learning from their environment and making informed decisions to achieve specific objectives. This research paper explores the potential of a reinforcement learning-based swarm of USVs to reduce marine debris in prominent water bodies while ensuring that inaccessible parts of larger water bodies are adequately cleaned by the collective strength of the swarm. The strategy involves collaboration between multiple USVs within a SWARM. The purpose of this research is to utilize multi-agent reinforcement learning to effectively clear marine debris with a greater degree of efficiency and further implement the research using ROS. The paper presents a comprehensive literature review and discusses the challenges and opportunities in implementing such swarms.\par 

The accumulation of marine debris is caused in part by littering, storm winds, and poor waste management. Approximately 80 percent of it originates from land-based sources. Marine debris can majorly include plastic objects such as beverage bottles, bottle caps, good wrappers and plastic straws. We intend to reduce the overall marine debris present in prominent water bodies through a swarm of USVs. This method will help to reduce the amount of time necessary for the space to be cleaned. By combining the collective strength of the USV SWARM, it would also be possible to clean inaccessible areas of larger water bodies effectively. We are currently looking at a USV SWARM to coordinate navigation and target tracking of the trash, which can significantly reduce the time required for cleaning up marine trash by focusing on efficiency.\par

The purpose of this paper is to address the problem of a swarm of unmanned surface vehicles (USVs) communicating to achieve a common objective. In our case, we are trying to clear marine residues from large bodies of water using a group of three to four autonomous USVs. To reduce the complexity of the problem, we first examine trash that is uniformly sized while utilizing USVs that are surface vehicles.\par

\section{Related Work}
A number of studies have investigated the use of RL to optimize USV performance. In Wang et al. [3], data-driven performance-prescribed reinforcement learning (DDP-PPRL) was used to control an unmanned surface vehicle. A reinforcement learning (RL)-based approach to achieving optimal tracking control of an unknown unmanned surface vehicle (USV) is presented in Wang et. Al [4]. Zhao et al. [5] optimized path following for an underactuated USV using smooth-convergent deep RL, whereas an auto-tuning environment for static obstacle avoidance in USVs was proposed by Guardeño et al. [9]. RL-based swarm control of UAVs has also been investigated. In aerial reconfigurable intelligent surfaces, Samir et al. [1] optimized the same using RL, and a distributed deep RL was proposed by Han et al. [6] for straight-path following and formation control of USVs. Kim et al. [13] presented a multi-agent deep RL approach for path planning in unmanned surface vehicles (USVs), and Wang et al. [19] proposed an adaptive and extendable control method for USV formations using distributed deep RL. RL has also been utilized in USVs for collision avoidance. Meyer et al. [14] proposed COLREG-compliant collision avoidance for USVs, whereas Na et al. [11] utilized bio-inspired collision avoidance with deep RL in swarm systems. Luis et al. [17] employed censored deep RL for Autonomous Surface Vehicles patrolling large water resources. In addition, swarm intelligence-based methods for USV control have been investigated in a number of studies. Zhao et al. [2] employed deep RL with random braking for USV formation and path-following control, whereas Xin et al. [7] suggested a greedy mechanism-based particle swarm optimization for USV path planning. Other studies have investigated the use of RL for USV path planning and localization. Yan et al. [10] proposed an RL-based method for AUV-assisted localization in the Internet of Underwater Things, whereas Yu et al. [21] developed a USV path planning method with velocity variation and global optimization using the AIS service platform. Lastly, a number of studies have surveyed UAVs, including their primary challenges and future trends. For example, Jorge et al. [8] provided an overview of USVs for disaster robotics, whereas Zereik et al. [18] analyzed the challenges and future trends in marine robotics. In conclusion, RL has been extensively utilized for USV control, and swarm control utilizing ROS is an active area of research. The studies cited in this section provide a comprehensive overview of the pertinent literature, including methods for optimizing USV performance, collision avoidance, and path planning, among others.\par

\section{Proposed Algorithm}
When we deal with the problem of cleaning up marine trash, we assume we are working in highly complex surroundings. As a result, the set of states and actions in the environment are complex and cannot be dealt with by the simple Deep Q-Network Architecture which is generally used to handle simple feedback control tasks. To overcome this problem, we look at a paradigm of reinforcement learning architectures known as Actor-Critic Networks. These architectures are well-known for their capability to handle continuous action spaces efficiently. The algorithm we use initially for the single agent environment is the Deep Deterministic Policy Gradient. This is an Actor-Critic Network for complex control environments. The algorithm uses a combination of target actor-critic and actor-critic networks to converge to an optimal solution by modifying the parameters of the actor-critic network using the target networks; these are copies of the value and policy networks used to calculate the target values within the training process. The variance of value estimates can be reduced by DDPG by periodically updating the target networks, further improving the learning process stability. The networks are updated based on a hyperparameter ($\tau$), which determines the rate at which updates are performed. Then we move to a multi-agent environment where we emphasize the MADDPG algorithm, which performs the same task of coordinated navigation in the SWARM. In a multi-agent SWARM environment, the robots now have a common goal to achieve rather than individual goals. Every USV of the SWARM takes actions that benefit the goal. DDPG has been successfully applied to a variety of tasks, including robotic control, navigation, and game-playing. For example, DDPG has been employed to train robot arms to perform reach and grasp tasks.\par

\subsection{Multi-Agent Reinforcement Learning}
In MARL, the paradigm for which is shown in Figure 1, numerous agents interact with one another and the environment to accomplish a common goal. Although each agent receives and acts upon separate observations of the environment, all agents partake in the rewards and goals. MARL is a potent method for complicated tasks requiring cooperation and coordination between numerous actors. In ROS, a SWARM of marine trash collection robots can be trained to work together to collect trash and clean up the environment using MARL. Every robot would get unique information about its surroundings, including the proximity to the trash and other robots, and act accordingly, such as heading towards the nearest piece of trash or avoiding collisions with other robots. All the robots would share in the benefits and goals, which would promote cooperation and coordination. By using the right algorithms and training techniques, such as Q-learning or actor-critic methods, and adjusting the rewards and objectives to incentivize cooperation and coordination among the robots, the MARL schema in the image can be used to solve the problem of SWARM marine trash collection in ROS. Moreover, strategies like decentralized control and communication can be used to boost the swarm's performance. Owing to the aforementioned factors, we believe a method based on MARL would promote coordination while optimizing the SWARM's purpose.\par

\begin{figure}[h]
\includegraphics[scale=0.35]{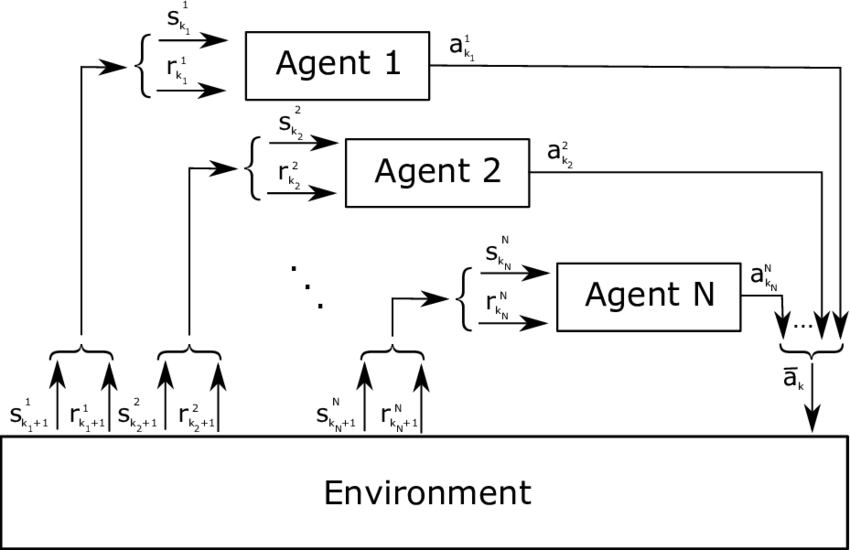}
\centering
\caption{Schematic of Multi-agent Reinforcement Learning Paradigm}
\label{fig:marl}
\end{figure}

\subsection{Deep Deterministic Policy Gradient}
The DDPG algorithm for a single marine trash-collecting robot employing DDPG in ROS is shown in a generalized form in Figure 2 below. A deterministic policy that links the agent's observations to its actions is learned for the agent by the actor-critic DDPG algorithm [22]. A robot used to gather marine trash might make observations about the area around trash and barriers, the water's depth, and the robot's own speed and heading.\par

The actor-network receives the agent's observations and produces a deterministic action. The output of the actor-network in the context of a marine-trash-collecting robot would be a velocity command for the robot, directing it towards the closest trash while avoiding obstacles. The critic network receives the agent's observations and actions as input and produces a value function ($Q_i$) that estimates the agent's expected cumulative reward from that state-action pair ($s_i$,$a_i$). This value function is used to assess the quality of the agent's actions and to update the actor's policy to maximize the expected cumulative reward R. The DDPG algorithm is an effective method for teaching a marine-trash collecting robot to navigate its environment, identify and collect trash, and avoid obstacles. Because the DDPG is suitable for continuous action areas, it is a good choice for controlling the movement of a marine-trash collection robot.\par

The swarm of marine trash collection robots can avoid overfitting to recent experiences by using experience replay. This can strengthen the learned policies and improve agent coordination in the task of marine trash collection.\par

To improve the stability and convergence of the learning process, target networks are used in the training of a single-agent marine-trash collection robot in ROS. The agent is equipped with two neural networks: the local network and the target network. During training, the local network is updated, while the target network is updated on a regular basis to match the weights of the local network. This helps to reduce the variance in Q-value estimates and stabilize neural network weight updates, resulting in better performance and higher rewards.\par

\begin{figure}[h]
\includegraphics[scale=0.45]{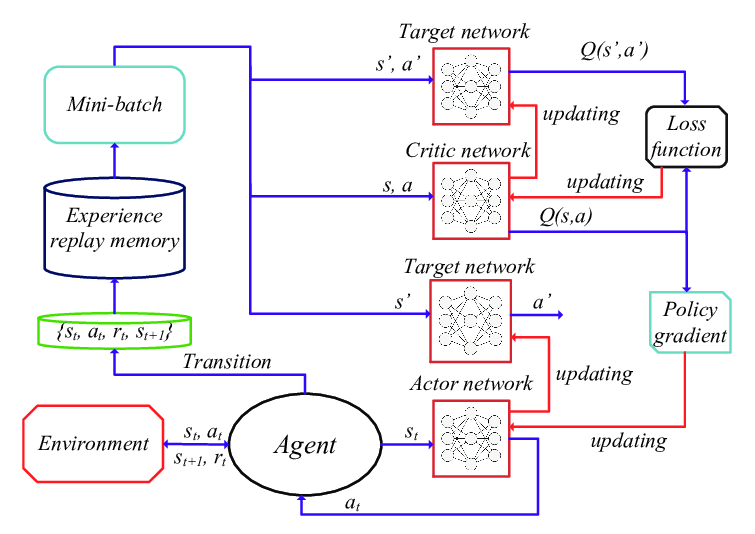}
\centering
\caption{Schematic of Deep Deterministic Policy Gradient}
\label{fig:ddpg}
\end{figure}

\vspace{1mm} 

\subsection{Multi-agent Deep Deterministic Policy Gradient}
The architecture of the MADDPG algorithm is depicted in Figure 3. This is a variant of the DDPG algorithm that is used in multi-agent systems to grasp decentralized policies for each agent in a cooperative or competitive setting [23]. It depicts two major components: a centralized critic network and decentralized actor networks. Each agent is related to its own actor-network, which receives its own observations as input and produces its own action. The centralized critic network takes all agents' joint states and actions as input and outputs a single value function that estimates the team's expected total reward.\par

\begin{figure}[H]
\includegraphics[scale=0.55]{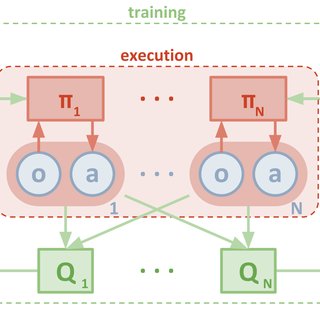}
\centering
\caption{Schematic of Multi-agent Deep Deterministic Policy Gradient}
\label{fig:maddpg}
\end{figure}

For The estimation of the Q-value for the action of each agent, further, the critic network uses a Q-network which allows the critic to estimate the total expected reward while taking into consideration, the actions of all agents in the system. The agent's observations from the environment are used by the actor networks to generate an action as an output, which is then combined with the actions of the other agents of the swarm and input into the critic network to estimate the Q-value. The critic network predicts the overall reward as the output, and the actor networks maximize it.

Each robot in the swarm could be thought of as an agent. The robots learn how to cooperate in order to properly collect the marine trash in the water body. The MADDPG algorithm is used here, to teach decentralized policies to each robot, enabling them to effectively work together and collect as much trash as possible.\par

The Q-values generated are used to update each agent's actor network during training. The critic network is updated based on the collective experience of all agents. MA-DDPG-based agents learn to interact without requiring contact with other agents. Target networks help in lowering variance in Q-value estimates and stabilizing neural network updates, leading to increased agent coordination and performance.\par

\subsection{Robot Operating System (ROS)}
To enhance the development process to build and control robotics devices, ROS is the most prominently used open-source platform. The MA-DDPG algorithm can be used to manage a swarm of marine trash-collecting robots. The MA-DDPG algorithm allows each robot in the swarm to design its own independent policy which helps them to boost the common reward as well as coordinate with the other robots of the swarm to achieve a shared goal. Here, ROS can be used as a platform for controlling the swarm and achieving the goal.\par  

The ROS paradigm of the publisher-subscriber model can be used to establish communication between the marine trash-collecting robots. It helps publishers send messages about a given topic and subscribers receive messages about the same topic. Each robot may have a publisher who sends messages about its current position, velocity, and other data required. After that, the robots in the swarm will subscribe to those signals to track the movements of the other robots. Based on the data, the robots collaborate and coordinate to collect waste by using the above ROS communication paradigm.\par

\begin{figure}[H]
\includegraphics[scale=0.24]{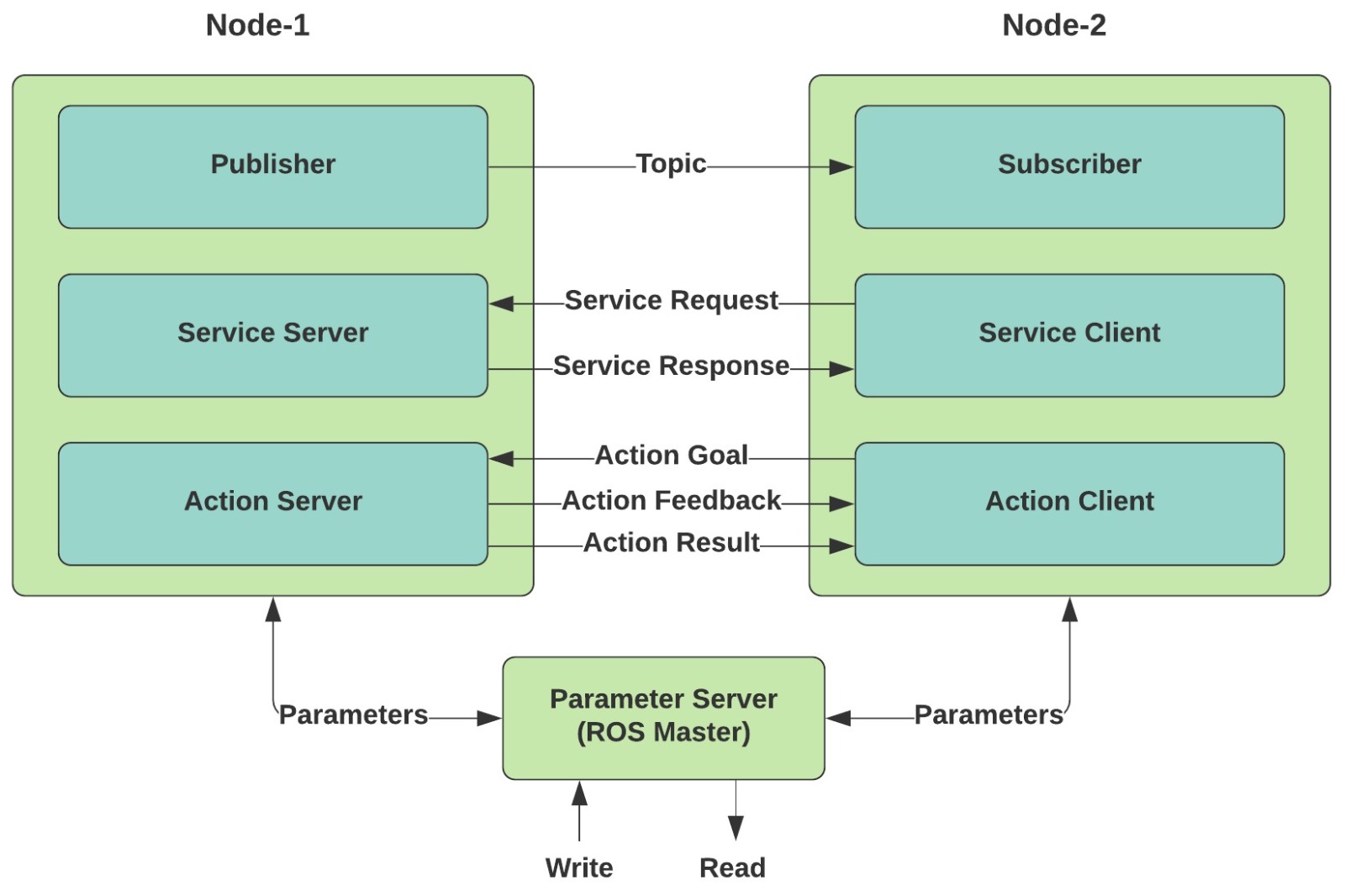}
\centering
\caption{Schematic of ROS Communication System}
\label{fig:algo}
\end{figure}

\begin{figure}[H]
\includegraphics[scale=0.55]{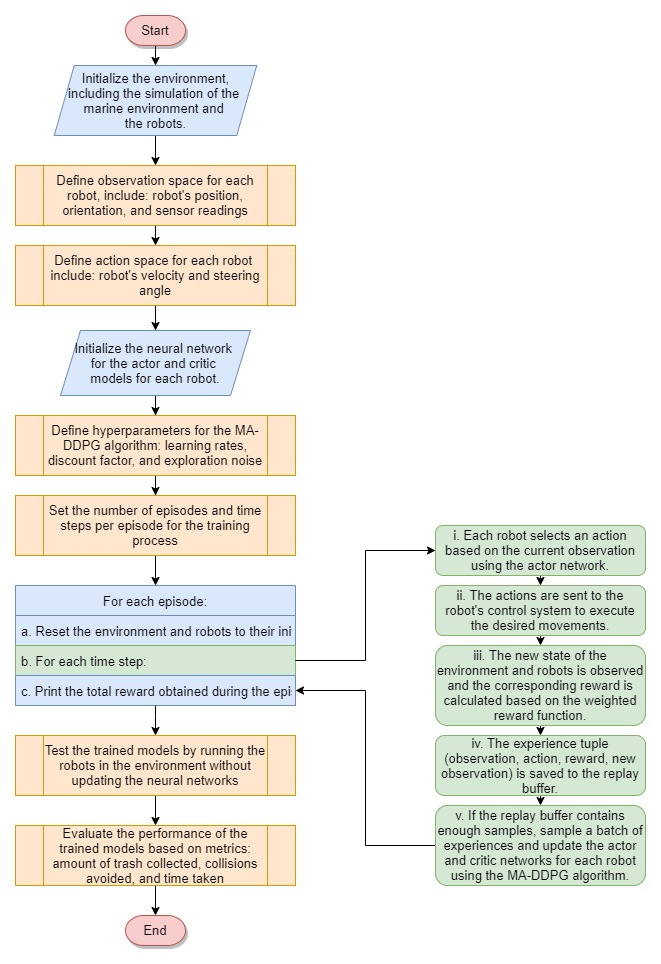}
\centering
\caption{Schematic of the Proposed Workflow}
\label{fig:algo}
\end{figure}

\subsection{Observation and Action Spaces}

To utilise MA-DDPG with ROS, each USV must have an observation space that enables it to examine its surroundings and take the appropriate actions required. The USV's position, orientation, left-right propeller linear and angular speeds, laser scan data, and true-false value for trash detection are stored within the observation space. The positional coordinates of the USV reveal its location within the water body. This data is essential for the USV to navigate and move toward the detected trash. The orientation of the USV, which conveys the overall direction in which it is facing, is very useful in determining how to progress toward the trash. The orientation of the robot provides information on the direction it is facing, which is useful for determining how to progress towards the trash. The propeller and angular speeds are utilized to regulate the robot's movement in the water. The robots may work together to travel towards and gather rubbish in a coordinated manner by modifying their speeds based on their observations. The distance from garbage calculated using laser scan data indicates the distance between the robot and the rubbish. This knowledge is critical for making decisions on how to move efficiently towards the trash. Finally, the trash detection boolean variable indicates whether or not the robot has identified any rubbish in its proximity. This information is critical for deciding how to approach and collect waste.\par

MA-DDPG is used to coordinate the robots' behaviours based on their observations, ensuring that they cooperate to fulfil their common aim of cleaning up the ocean. The agents learn to conduct activities that maximize a reward signal, which is intended to encourage the robots to travel towards and collect rubbish as efficiently as possible. The agents learn to coordinate their actions and operate as a team to achieve their goal of cleaning up the ocean through training. Propeller speeds are employed as the action space for each robot in the ROS challenge of a swarm of marine trash collection robots using MA-DDPG. The propeller speeds regulate the movement of the robots in the water, allowing them to travel towards and collect rubbish efficiently. The robots may work together to collect rubbish in a coordinated and efficient manner by modifying their propeller speeds based on their observations.MADDPG is used to coordinate the robots' actions, ensuring that they all work together to clean up the water body.\par

\section{Implementation}

\subsection{Architecture of WAM-V in ROS and Gazebo}

To solve the challenge at hand, we have made use of the Wave Adaptive Modular Vessel (WAM-V) which has been used for a variety of missions in real-time such as oceanographic data collection, environmental monitoring, and maritime security. To simulate and control the WAM-V in a virtual environment, we utilise the Robot Operating System (ROS) and Gazebo.\par

\begin{figure}[H]
\includegraphics[scale=0.23]{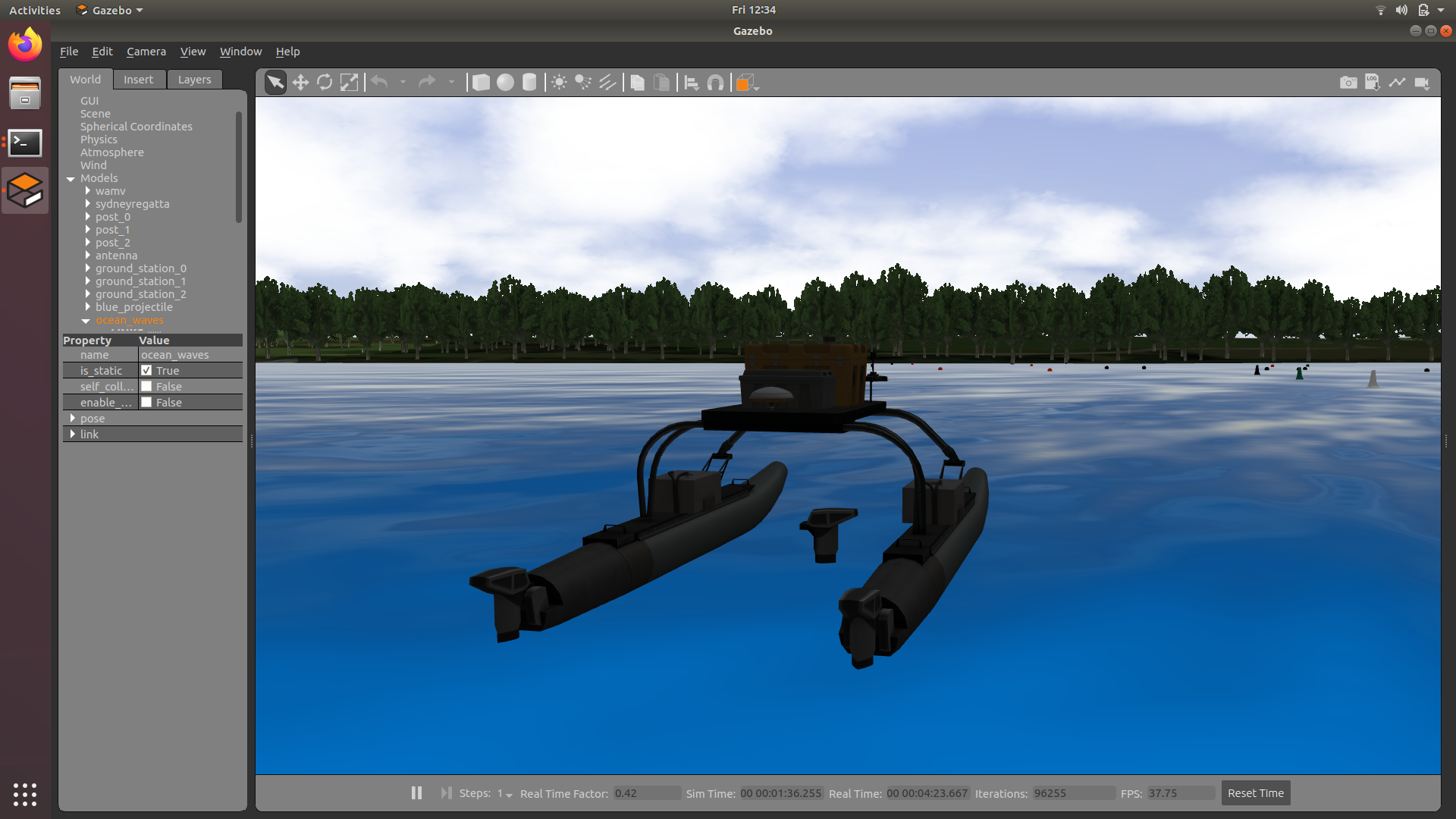}
\centering
\caption{WAM-V in Gazebo}
\label{fig:algo}
\end{figure}

The distinct nodes present within the architecture of the WAM-V in ROS and Gazebo, each of which is accountable for a certain task such as sensor data processing, motion control, and communication with external devices.\par

\vspace{3cm}

The data from a variety of sensors is collected by the sensor nodes, including GPS, IMU and Depth Sensors, and is transmitted to the data processing nodes. This data is analyzed by these nodes to provide the position, orientation, and environment information of the WAM-V.\par

\vspace{1.9cm}

\begin{figure}[H]
\raggedright
\includegraphics[scale=0.34]{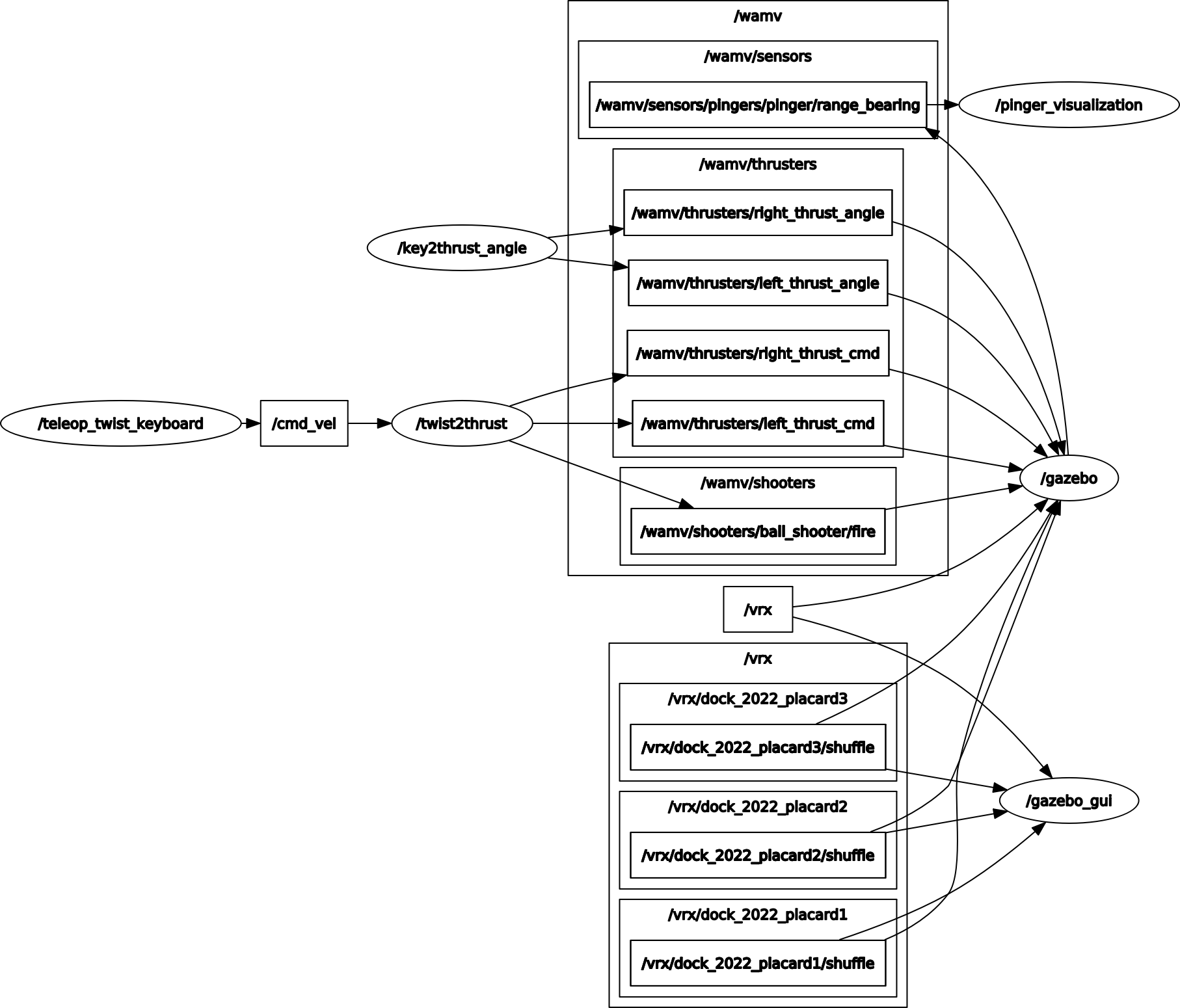}
\caption{RQT graph of WAM-V}
\label{fig:algo}
\end{figure}

To visualise the data, the RViz tool can be used which renders the WAM-V and its surroundings in real time. This tool can visualise the sensor data in a meaningful manner, providing insight into the WAM-V's operation and surrounding environment.\par

\begin{figure}[H]
\includegraphics[scale=0.23]{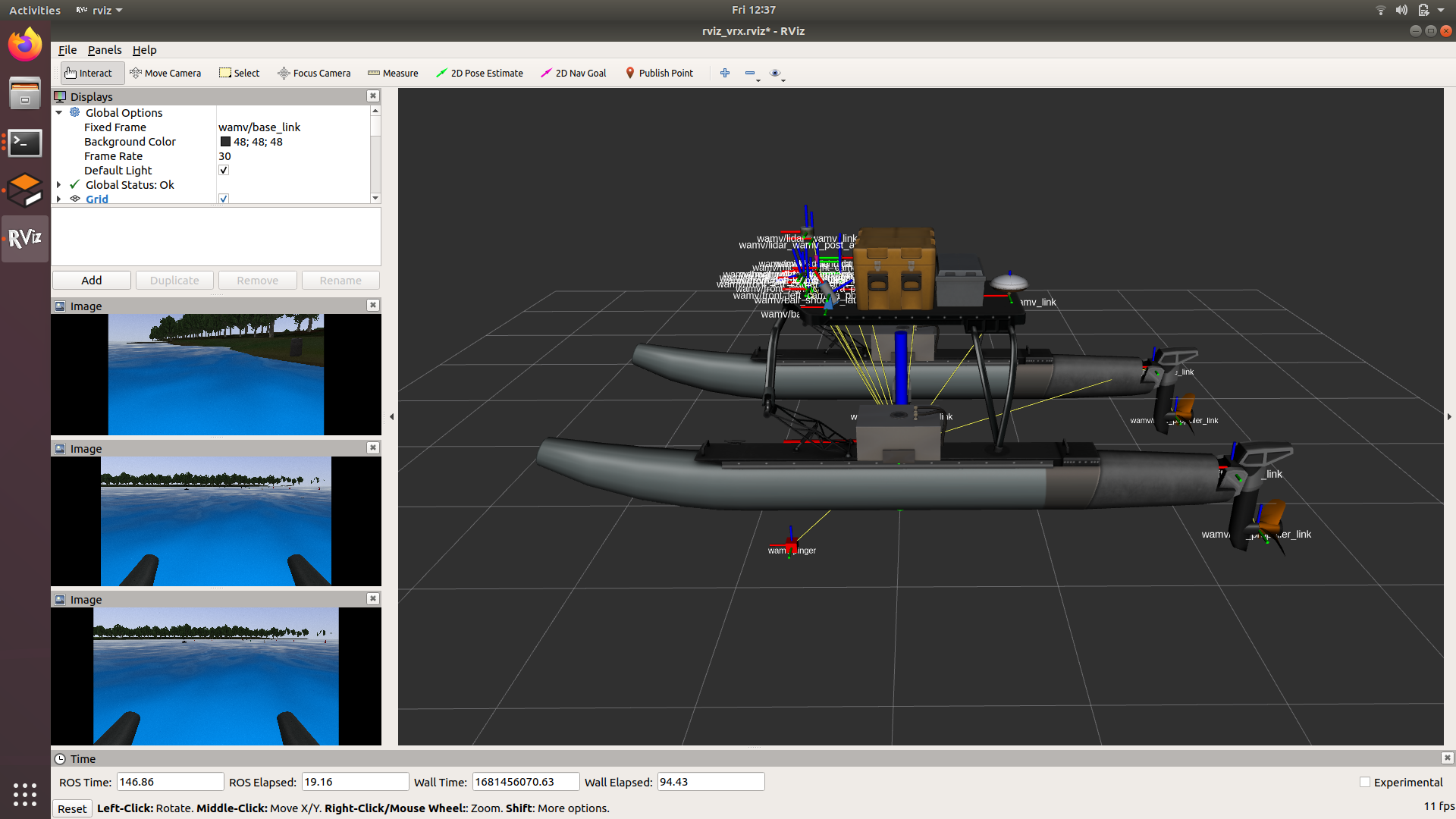}
\centering
\caption{WAM-V in RViz}
\label{fig:algo}
\end{figure}

Additionally, the RViz tool can interact with the WAM-V allowing for the testing and evaluation of the control algorithms. For instance, the movement of the right and left propellers of the WAM-V can be controlled by sending commands from the RViz tool.\par

\begin{figure}[H]
\includegraphics[scale=0.23]{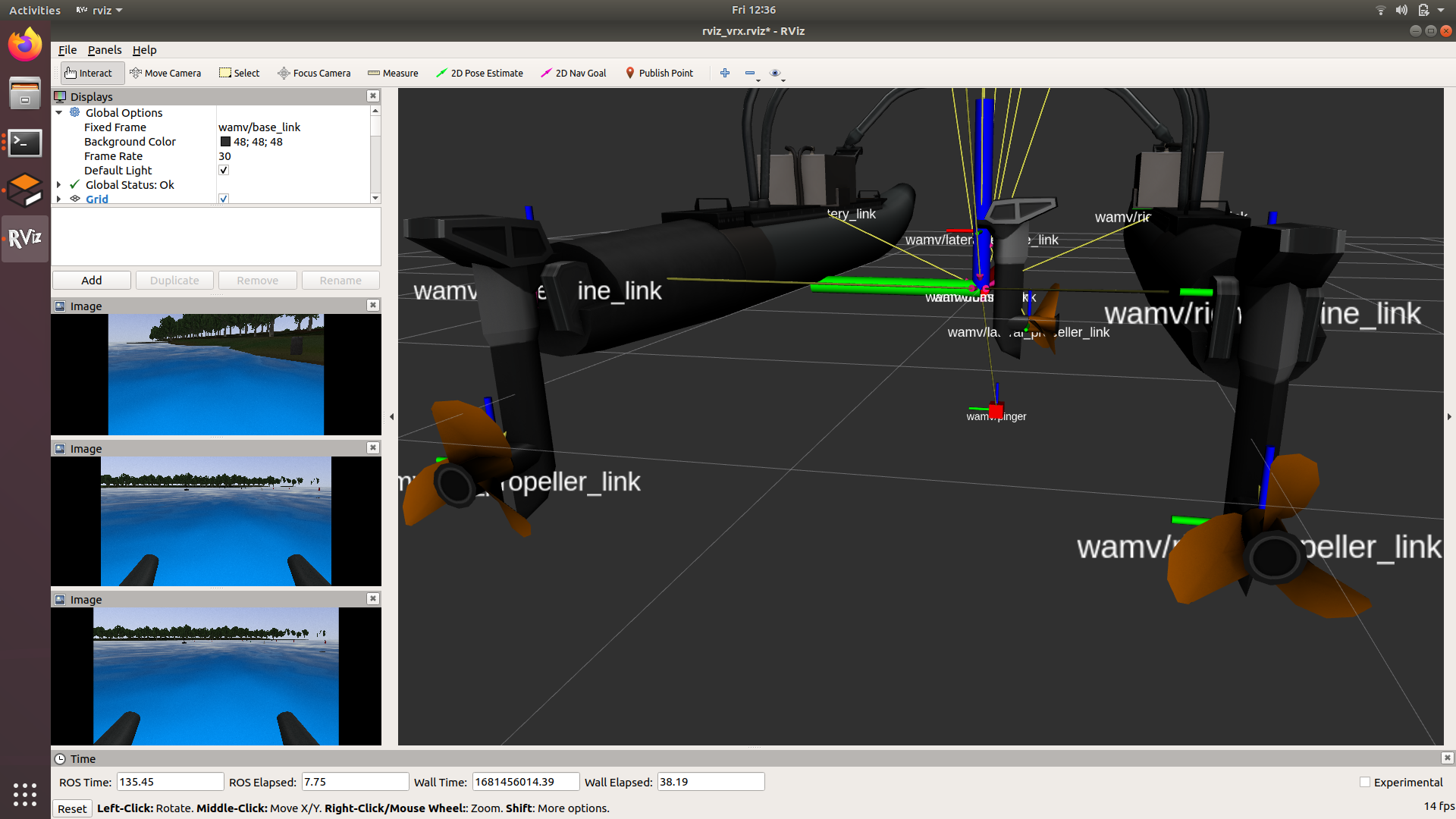}
\centering
\caption{WAM-V propeller view in RViz}
\label{fig:algo}
\end{figure}

Implementation of the proposed algorithm, with WAM-V, in ROS and Gazebo provides a versatile platform for acquiring, processing, and visualising sensor data. As mentioned before, the utilisation of the RViz tool provides real-time visualisation and interaction with the WAM-V.\par

\subsection{Overview of the openai\textunderscore ros package for Integrating ROS and OpenAI}

To implement the aforementioned algorithms within ROS, we make use of the openai\textunderscore ros package. The package contains a collection of components that offer to interface with OpenAI Gym. The nodes facilitate the exchange of data between ROS and OpenAI Gym, allowing ROS-based robotics applications to be integrated with OpenAI's reinforcement learning algorithms.\par

To utilise the package, a training environment is created which represents the robot application to be trained.\par

\begin{figure}[H]
\includegraphics[scale=0.4]{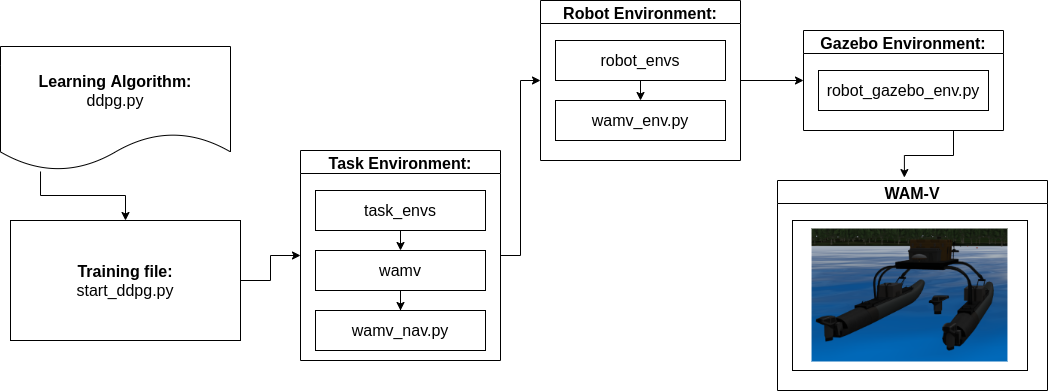}
\centering
\caption{openai\textunderscore ros environment with WAM-V}
\label{fig:algo}
\end{figure}

States, actions, and rewards are comprised within the Training environment which is further implemented as a ROS node that communicates with the openai\textunderscore ros package's OpenAI Gym interface. After establishing the training environment, a corresponding training script is composed which utilises OpenAI's reinforcement learning algorithms to train a policy for the given environment. Using the openai\textunderscore ros package, the training script is enabled to interact with the training environment to acquire data for training the policy. During training, the reinforcement learning algorithm interacts with the training environment and receives data on the current state and further chooses actions based on the current policy. The environment then offers the algorithm a reward based on the action performed, and the algorithm modifies its policy accordingly.\par

\section{Reward Shaping}
We devise a weighted reward function $R$ for controlling a SWARM of marine trash-collecting robots, based on the problem's objectives and limitations. The essential goal of the problem is to collect as much trash as possible while minimizing collisions.

\textbf{\textit{Reward for collection ($R_{collect}$)}}:  Reward is given to the robots when the trash is collected.
\\\textbf{\textit{Penalty for collision ($P_{coll}$)}}: Negative reward is given to a robot when it collides with another robot or an obstacle.
\\\textbf{\textit{Penalty for Time Taken ($P_{time}$)}}: Negative reward is given to the robots as time passes by to incentivize a quicker collection of trash.
\\\textbf{\textit{Reward for coordination ($R_{coord}$)}}: The coordination reward $R_{coord}$ is a positive reward for the robots to make them cooperate with each other and gather the trash. This could be accomplished by rewarding the robots for collecting trash in closer proximity  to each other. 

The above-weighted reward function provides positive rewards for collecting trash and negative rewards for delays caused and collisions occurred. Additionally, a reward is given to the robots based on the proximity of the robots to one another and thus the coordination achieved between them. The reward function can be written as:
\vspace{2.5mm} 

\begin{center}
$R$ = $w_1$* $R_{collect}$ - $w_2$*$P_{coll}$ - $w_3$* $P_{time}$ + $w_4$*$R_{coord}$
\end{center}

\vspace{2.5mm} 

Here, $w_1$, $w_2$, and $w_3$ represent the weights that have been assigned to each reward term. The collection reward can be defined as a function of the amount of garbage the robots collect together, whereas the collision penalty can be defined as a negative reward for each collision the robots cause. In a similar fashion, the time penalty can be defined as a negative reward for the summation of each of the time steps the robots take. The weights assigned to each term can be adjusted to control the behaviour of the swarm of marine trash-collecting robots. A high weight on the collection reward term would motivate the robots to prioritize trash collection, whereas a high weight on the collision penalty term would motivate them to avoid collisions. In conclusion, the previously formulated weighted reward function can be used to incentivize robots to accomplish the primary goal while avoiding collisions and time delays.

The weight allocated to the coordination reward term is denoted by $w_4$. The sum of all pairs of robots $(i, j)$ is calculated, and $D(i,j)$ denotes the distance between the two robots. The $(1 - D(i,j))$ term ensures that the robots are compensated for their proximity to one another. A larger weight on the reward for coordination would motivate the robots to coordinate and collaborate with each other in order to collect trash more efficiently.

Here, The coordination reward is defined as the product of the weight applied to the coordination reward term ($w_4$) and $(1 - D(i,j))$ over all pairs of robots $(i, j)$. This term ensures that the robots are rewarded for their closeness in distance to one another. Hence, the coordination reward incentivizes collaboration and coordination among the swarm of robots to achieve the primary goal of collecting as much trash as possible while minimizing collisions with one another and the obstacles in the way and reducing the time delays.

\section{Applications}
ROS autonomous unmanned vehicle swarms based on Reinforcement Learning (RL) have several possible applications. Firstly there is environmental monitoring, wherein an RL-based swarm of autonomous unmanned vehicles in ROS can be used for environmental monitoring and surveillance. The swarm can be taught to navigate through many terrains and environments, record data, and communicate with the base station. Next, there is search and rescue. This is where an RL-based swarm of autonomous unmanned vehicles in ROS can be used for search and rescue operations in emergency circumstances. In disaster-stricken areas, a trained swarm of USVs can be used to search for survivors and provide the rescue team with real-time data. The third application is within agriculture. For precision agriculture, a swarm of autonomous unmanned vehicles based on RL and powered by ROS can be utilised. The swarm can be taught to monitor crops, locate disease outbreaks, and apply required fertilisers or pesticides. ROS autonomous unmanned vehicles also provide support in the transportation sector. An RL-based swarm of autonomous unmanned vehicles in ROS can be used for transference. The swarm can be taught to transport goods or individuals from one location to another in a safe and efficient manner. Next, it aids in Infrastructure inspection where an RL-based swarm of autonomous unmanned vehicles in ROS can be used to inspect essential infrastructures such as bridges, pipelines, and power lines. The swarm can be trained to detect liabilities and provide early warning indicators of potential failures. Lastly, it assists in military applications where An RL-based swarm of autonomous unmanned vehicles in ROS can be used for military reconnaissance and surveillance. The swarm can be trained to conduct covert operations and gather intelligence in hostile territories.


\section{Conclusion}
The MA-DDPG-based SWARM of Autonomous Unmanned Vehicles in ROS provides a flexible and robust solution for decentralised swarm control. The MA-DDPG algorithm gives each and every agent the ability to learn from its own experiences and interactions with other agents, therefore enhancing its scalability and adaptability. The integration of RL with ROS supplies a modular and malleable surface for developing and integrating the many different components of the swarm. The simulation environment used in this study allows us to evaluate the execution of the proposed system in numerous situations and compare it with other existing approaches. In terms of scalability, adaptability, and efficiency, the proposed MA-DDPG-based swarm outperforms all other existing procedures. The swarm was able to navigate through complex environments and avoid obstacles while coordinating with other agents all while maintaining stable communication. Despite the encouraging results, there are still challenges that need to be fully addressed in order to realise the full potential of the MA-DDPG-based swarm of autonomous unmanned vehicles in ROS. These challenges include improving the learning efficiency of the agents, addressing and resolving safety concerns, and making sure of the ethical use of autonomous systems.\par

The proposed MA-DDPG-based swarm of autonomous unmanned vehicles in ROS represents an innovative and promising strategy for the development of intelligent, self-aware and autonomous systems. Integration of RL with ROS opens up exciting prospects for future research and development in the field.\par


\end{document}